\def\year{2019}\relax
\documentclass[letterpaper]{article} %DO NOT CHANGE THIS
\usepackage{aaai19}  %Required
\usepackage{times}  %Required
\usepackage{helvet}  %Required
\usepackage{courier}  %Required
\usepackage{url}  %Required
\usepackage{graphicx}  %Required
\usepackage{latexsym}
\usepackage{multirow}
\usepackage{arydshln}
\usepackage{amsmath, amssymb}
\usepackage{bm}
\usepackage{pifont}
\newcommand{\cmark}{\ding{51}}
\newcommand{\LF}[1]{\ensuremath{\texttt{#1}}}
\newcommand{\tto}{\!\to\!}
\newcommand{\citet}[1]{\citeauthor{#1}~\shortcite{#1}}
\newcommand{\citealp}[1]{\citeauthor{#1}~\citeyear{#1}}

\title{Combining Axiom Injection and Knowledge Base Completion for Efficient Natural Language Inference}

\author{Masashi Yoshikawa\\
Nara Institute of Science and Technology, Nara, Japan\\
yoshikawa.masashi.yh8@is.naist.jp\\
\And
Koji Mineshima\\
Ochanomizu University, Tokyo, Japan\\
mineshima.koji@ocha.ac.jp\\
\AND
Hiroshi Noji\\
Artificial Intelligence Research Center, AIST, Tokyo, Japan\\
hiroshi.noji@aist.go.jp\\
\And
Daisuke Bekki\\
Ochanomizu University, Tokyo, Japan\\
bekki@is.ocha.ac.jp\\
}

\begin{document}
\maketitle
\begin{abstract}
    In logic-based approaches to reasoning tasks such as Recognizing Textual Entailment~(RTE),
    it is important for a system to have a large amount of knowledge data.
    However, there is a tradeoff between
    adding more knowledge data for improved RTE performance
    and maintaining an efficient RTE system, as such a big database is problematic
    in terms of the memory usage and computational complexity.
    In this work, we show the processing time of a state-of-the-art logic-based RTE system can be significantly
    reduced by replacing its search-based axiom injection~(abduction) mechanism by that based on Knowledge Base Completion~(KBC).
    We integrate this mechanism in a Coq plugin that provides a proof automation tactic for natural language inference.
    Additionally, we show empirically that adding new knowledge data contributes to better RTE performance
    while not harming the processing speed in this framework.
\end{abstract}

\section{Introduction}
RTE is a challenging NLP task where the objective is
to judge whether a hypothesis $H$ logically follows from premise(s) $P$.
Advances in RTE have positive implications in other areas such as
information retrieval, question answering and reading comprehension.
Various approaches have been proposed to the RTE problem in the literature.
Some methods are based on deep neural networks~\cite{Rocktschel2015ReasoningAE,P18-1224,nie2017shortcut}, where
a classifier is trained to predict using $H$ and $P$ encoded in a high-dimensional space.
Other methods are purely symbolic~\cite{Bos:2004:WSR:1220355.1220535,mineshima-EtAl:2015:EMNLP,abzianidze:2015:EMNLP}, where logical formulas
that represent $H$ and $P$ are constructed and used in a formal proof system.
In this paper, we adopt a strategy based on logic, encouraged by the high-performance that
these systems achieve in
linguistically complex datasets~\cite{mineshima-EtAl:2015:EMNLP,abzianidze:2015:EMNLP},
which contain a variety of semantic phenomena that are still challenging for
the current neural models~\cite{DBLP:journals/corr/abs-1804-07461}.

Contrary to the end-to-end machine learning approaches, a logic-based system must explicitly maintain lexical knowledge necessary for inference. A critical challenge here is to deal with such knowledge in an efficient and scalable way.
A promising approach in past work is on-demand axiom injection (abduction mechanism; \citealp{martinezgomez-EtAl:2017:EACLlong}), which allows one to construct knowledge between words in $P$ and $H$ as lexical axioms, and feed them to a logic prover when necessary.
Combined with ccg2lambda \cite{mineshima-EtAl:2015:EMNLP}, a higher-order logic-based system, their method demonstrates that injecting lexical knowledge from WordNet \cite{Miller:1995:WLD:219717.219748} and VerbOcean \cite{chklovski-pantel:2004:EMNLP} significantly improves the performance.

Although their method provides a basis for handling external knowledge with a logic-based RTE system,
there still remains a practical issue in terms of scalability and efficiency.
Their abduction mechanism generates relevant axioms on-the-fly for a present $P$ and $H$ pair, but this means we have to maintain a large knowledge base inside the system.
This is costly in terms of memory, and it also leads to slower search due to a huge search space during inference.
WordNet already contains relations among more than 150,000 words, and in practice, we want to increase the coverage of external knowledge more by adding different kinds of database such as Freebase.
To achieve such a scalable RTE system, we need a more efficient way to preserve database knowledge.

In this paper, we present an approach to axiom injection, which, by not holding databases explicitly, allows handling of massive amount of knowledge without losing efficiency.
Our work is built on Knowledge Base Completion~(KBC), which recently has seen a remarkable advancement in the machine learning community.
Although KBC models and logic-based approaches to RTE have been studied separately so far, we show that they can be combined to improve the overall performance of RTE systems.
Specifically, we replace the search of necessary knowledge on the database with a judgment of whether the triplet $(s, r, o)$ is a fact or not in an $n$-dimensional vector space that encodes entities $s$ and $o$ and relation $r$.
For each triplet, this computation is efficient and can be done in $O(n)$ complexity.
To this end we construct a new dataset from WordNet for training a KBC model that is suitable for RTE.
We then show that this approach allows adding new knowledge from VerbOcean without losing efficiency.

Throughout this paper, we will focus on inferences that require lexical knowledge such as synonym and antonym and its interaction with the logical and linguistic structure of a sentence, distinguishing them from common sense reasoning (e.g., \textit{John jumped into the lake} entails \textit{John is wet}) and inferences based on world knowledge (e.g., \textit{Chris lives in Hawaii} entails \textit{Chris lives in USA}).
For evaluation, we use the SICK (Sentences Involving Compositional Knowldedge) dataset~\cite{MARELLI14.363.L14-1314}, which focuses on lexical inferences combined with linguistic phenomena.\footnote{
Large-scale datasets for training neural natural language inference models such as
SNLI~\cite{snli:emnlp2015} and MultiNLI~\cite{N18-1101}
are not constrained to focus on lexical and linguistic aspects of inferences,
which can produce confounding factors in analysis, hence are not suitable for our purposes.
}

Another advantage of our approach is that we can complement the missing lexical knowledge in existing knowledge bases as latent knowledge.
The previous method is limited in that it can only extract 
relations that are directly connected or reachable
by devising some relation path (e.g. transitive closure for hypernym relation);
however, there are also lexical relations that are not explicitly available and hence latent in the networks.
To carefully evaluate this aspect of our method, we manually create a small new RTE dataset, where each example requires complex lexical reasoning, and find that our system is able to find and utilize such latent knowledge that cannot be reached by the existing approach.

Our final system achieves a competitive RTE performance with \citeauthor{martinezgomez-EtAl:2017:EACLlong}'s one, while keeping the processing speed of the baseline method that does not use any external resources.
The last key technique for this efficiency is a new {\tt abduction} tactic,
a plugin for a theorem prover Coq~\cite{coq}.
One bottleneck of \citeauthor{martinezgomez-EtAl:2017:EACLlong}'s approach is that in their system
Coq must be rerun each time new axioms are added.
To remedy this overhead we develop {\tt abduction} tactic that enables searching knowledge bases and executing a KBC scoring function during running Coq.

Our contributions are summarized as follow:\footnote{
\label{codebase}
  All the programs and resources used in this work are publicly available at:
  \url{https://github.com/masashi-y/abduction_kbc}.
}
\begin{itemize}
\item We propose to combine KBC with a logic-based RTE system for efficient and scalable reasoning.
\item We develop an efficient abduction plugin for Coq, which we make publicly available.
\item We show that our techniques achieve a competitive score to the existing abduction technique while maintaining the efficiency of the baseline with no knowledge bases.
\item We construct a set of lexically challenging RTE problems and conduct extensive experiments to evaluate the latent knowledge our KBC model has learned.
We demonstrate many examples of those knowledge that are not available for the previous method.
\end{itemize}

\section{Related work}
\label{sec:relwork}

\subsection{Logic-based RTE systems}
\label{ssec:rtesys}

Earlier work on logic-based approaches to RTE
exploited off-the-shelf first-order reasoning tools (theorem provers and model-builders)
for the inference component~\cite{bos2005recognising}.
Such a logic-based system tends to have high precision and low recall for RTE tasks,
suffering from the lack of an efficient method to integrate external knowledge in the inference system.

Meanwhile, researchers in Natural Logic~\cite{van2007brief}
have observed that
the iteration depth of logical operators such as negations and quantifiers in natural languages is limited
and, accordingly, have developed a variety of proof systems
such as monotonicity calculus~\cite{icard2014recent}
adapted for natural languages that use small parts of first-order logic.

The idea of natural logic has recently
led to a renewed interest in symbolic approaches
to modeling natural language inference in the context of NLP~\cite{MacCartney-Manning08}.
In particular, theorem provers designed for natural languages have been recently developed,
where a proof system such as
analytic tableau \cite{abzianidze:2015:EMNLP}
and natural deduction \cite{mineshima-EtAl:2015:EMNLP}
is used in combination with wide-coverage parsers.
These systems allow a controlled use of higher-order logic,
following the tradition of formal semantics~\cite{Montague74},
and thereby have achieved efficient reasoning for logically complex RTE problems such as
those in the FraCaS test suite~\cite{cooper1994fracas}.
However, it has remained unclear how one can add robust external knowledge to
such logic-based RTE systems without loss of efficiency of reasoning.
The aim of this paper is to address this issue.

We use Coq,
an interactive proof assistant based on the Calculus of Inductive Constructions (CiC), 
in order to implement an RTE system with our method of axiom insertion.
Although Coq is known as an interactive proof assistant,
it has a powerful proof automation facility
where one can introduce user-defined proof strategies called \textit{tactics}.
The work closest to our approach in this respect
is a system based on Modern Type Theory~\cite{chatzikyriakidis2014natural}, which uses Coq as an automated theorem prover 
for natural language inferences.
It was shown that the system achieved high accuracy on the FraCaS test suite~\cite{bernardy17}.
However, the system relies on a hand-written grammar, suffering from scalability issues.
Additionally, this work did not evaluate the efficiency of theorem proving for RTE,
nor address the issue of how to extend their RTE system with robust external knowledge.

Generally speaking, it seems fair to say that
the issue of efficiency of logic-based reasoning systems 
with a large database has been underappreciated in the literature on RTE.
To fill this gap, we investigate how our logic-based approach to RTE,
combined with machine learning-based Knowledge Base Completion,
can contribute to robust and efficient natural language reasoning.

\begin{figure*}
    \centering
    \includegraphics[scale=0.54]{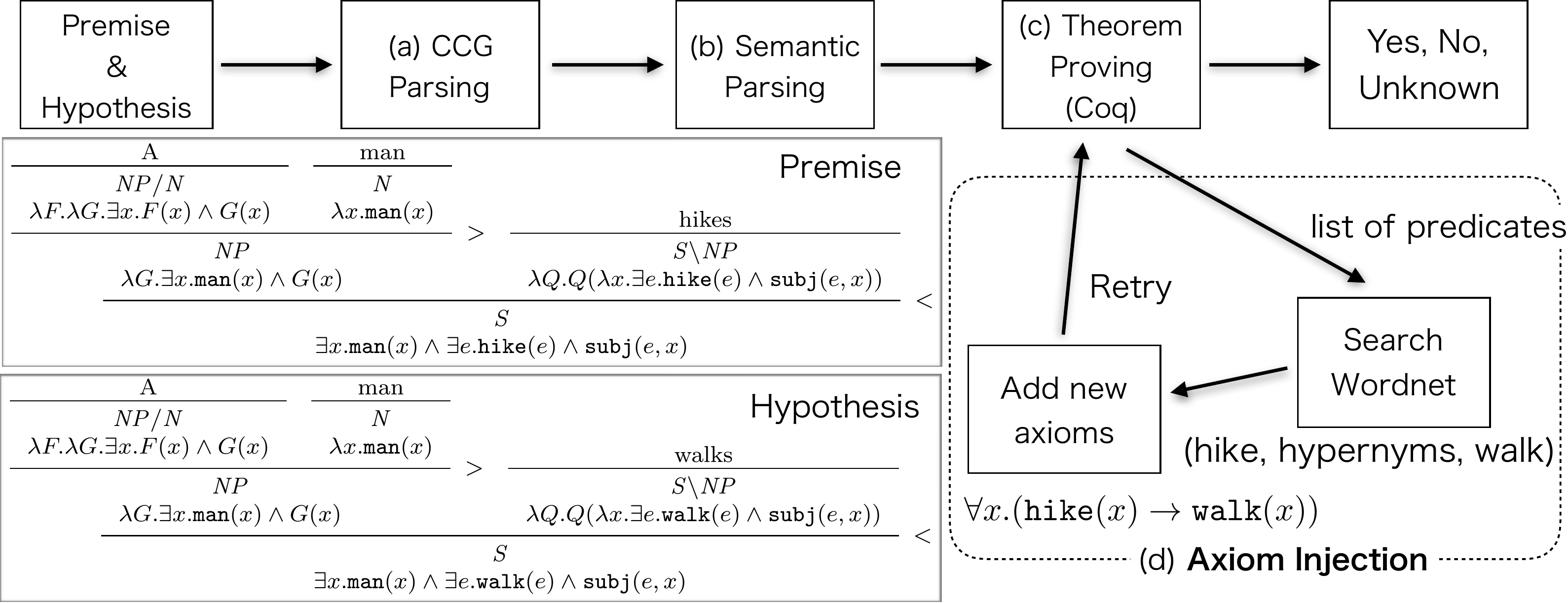}
    \caption{A pipeline of ccg2lambda.
    It firstly applies CCG parser to premise ($P$) and hypothesis ($H$) sentences~(a),
    and then convert them to logical formulas~(b).
    It tries to prove if entailment (contradiction) can be established
    by applying Coq to the theorem $P \to H$ ($P \to \lnot H$)~(c).
    If the proving fails, it tries axiom injection~(d).
    }
    \label{ccg2lambda}
\end{figure*}

\subsection{Knowledge Base Completion~(KBC)}
\label{sec:kbc}
Several KBC models have been proposed in the literature~\cite{Bordes:2013:TEM:2999792.2999923,trouillon2016complex,DBLP:journals/corr/DettmersMSR17}.
Among them we use ComplEx~\cite{trouillon2016complex}, which models triplet $(s,r,o)$ for entities $s, o \in \mathcal{E}$ and relation $r \in \mathcal{R}$ in an $n$-dimensional complex vector space\footnote{
    $\mathbb{C}$ denotes the set of complex numbers.
    For $x \in \mathbb{C}$, $Re(x)$ denotes its real part and $\overline{x}$ its complex conjugate.}:
\begin{align}
    \psi_r(s,o) = \sigma(Re(\langle {\bm e}_s, {\bm e}_r, \overline{{\bm e}_o} \rangle)), \label{complex}
\end{align}
where ${\bm e}_s, {\bm e}_r, {\bm e}_o \in \mathbb{C}^n$,
$\langle {\bm x},{\bm y},{\bm z} \rangle = \sum_i x_i y_i z_i$, and $\sigma$ is the sigmoid function.
Since Eq.~\ref{complex} consists of one dot-product among three vectors,
its computational complexity is $O(n)$.
The training is done by minimizing the logistic loss:

\begin{align}
    \sum_{((s,r,o), t) \in \mathcal{D}} t \log \psi_r(s,o) 
    + & (1 - t) \log (1 - \psi_r(s,o)), \label{logistic}
\end{align}
where $\mathcal{D}$ is the training data.
We have $t=1$ if the triplet is a fact and $t=0$ if not (negative example).
While negative examples are usually collected by negative sampling,
1-N scoring has been proposed to accelerate training~\cite{DBLP:journals/corr/DettmersMSR17}.
In 1-N scoring, unlike other KBC models that take an entity
pair and a relation as a triplet $(s, r, o)$ and score it (1-1 scoring),
one takes one $(s, r)$ pair and scores it against all entities $o \in \mathcal{E}$ simultaneously.
It is reported that this brings over 300 times faster computation of the loss for their convolution-based model.
This method is applicable to other models including ComplEx and
scales to large knowledge bases.
We employ this technique in our experiments.

\section{System overview}
\label{sec:ccg2lambda}

We build our system on ccg2lambda~\cite{mineshima-EtAl:2015:EMNLP}\footnote{\url{https://github.com/mynlp/ccg2lambda}}, an RTE system with higher-order logic. Figure~\ref{ccg2lambda} shows a pipeline of the system.

Note that although we use a particular RTE system to test our hypotheses, other logic-based RTE systems can also benefit from our KBC-based method. In so far as a lexical relation is modeled as a triplet, it could be adapted to their inference modules; for instance, if $r$ is a hypernym relation, a triplet $(s, r, o)$ is mapped to $s \sqsubseteq o$, where $\sqsubseteq$ is a containment relation in Natural Logic~\cite{MacCartney-Manning08} or a subtyping relation in Modern Type Theory~\cite{chatzikyriakidis2014natural}, rather than to $\forall x. (s(x) \to o(x))$ as in the standard logic we use in this paper.

\subsection{CCG and semantic parsing}
\label{ssec:ccg}

The system processes premise(s)~($P$) and a hypothesis~($H$) using
Combinatory Categorial Grammar~(CCG; \citealp{steedman:syntactic_process}) parsers.
CCG is a lexicalized grammar that provides syntactic structures transparent to semantic representations~(Figure~\ref{ccg2lambda}a).
In CCG, each lexical item is assigned a pair $(C, M)$ of the syntactic category $C$ and a meaning representation $M$ encoded as a $\lambda$-term;
for instance, ``man'' is assigned $(N, \lambda x. \LF{man}(x))$.
The parses (called \textit{derivation trees}) are converted into logical formulas by composing $\lambda$-terms assigned to each terminal word in accordance with combinatory rules~(Figure~\ref{ccg2lambda}b).

For the assignment of $\lambda$-terms, we use a template-based procedure,
where closed-class words (logical or functional expressions) are mapped 
to their specific meaning representations
and other words to schematic meaning representations based on CCG categories.
In this work, we adopt a semantic template based on Neo-Davidsonian Event Semantics~\cite{parsons:event}, where a sentence is mapped to a formula involving
quantification over events and a verb is analyzed as a 1-place predicate
over events using auxiliary predicates for semantic roles such as
\LF{subj} (see the formulas in Figure~\ref{ccg2lambda}b).
One of the main attractions of this approach is that it facilitates
the simple representation
of lexical relations for nouns and verbs, since both can be uniformly analyzed
as 1-place predicates (see the axioms in Table~\ref{generated_axioms}).

\subsection{Theorem proving}
\label{ssec:proving}

The system uses automated theorem proving in Coq~(Figure~\ref{ccg2lambda}c)
to judge whether entailment ($P \to H$) or contradiction ($P \to \lnot H$)
holds between the premise and the hypothesis.
It implements a specialized prover for higher-order features in natural language, which
is combined with Coq's build-in efficient first-order inference mechanism.
Coq has a language called Ltac for user-defined automated tactics~\cite{10.1007/3-540-44404-1_7}.
The additional axioms and tactics specialized for natural language constructions are written in Ltac.
We run Coq in a fully automated way,
by feeding to its interactive mode a set of predefined tactics combined
with user-defined proof-search tactics.

\subsection{Axiom insertion (abduction)}
\label{ssec:axiominsertion}

Previous work~\cite{martinezgomez-EtAl:2017:EACLlong} extends ccg2lambda with
an axiom injection mechanism using databases such as
WordNet~\cite{Miller:1995:WLD:219717.219748}.
When a proof of $T \to H$ or $T \to \neg H$ fails,
it searches these databases for
lexical relations that can be used to complete the theorem at issue.
It then restarts a proof search after declaring the lexical relations as axioms~(Figure~\ref{ccg2lambda}d).
This mechanism of on-demand insertion of axioms is called \textit{abduction}.

A problem here is that this abduction mechanism slows down the overall processing speed.
This is mainly due to the following reasons:
(1) searches for some sort of relations incur multi-step inference over triplets~(e.g. transitive closure of hypernym relation);
(2) the theorem proving must be done all over again to run an external program that searches the knowledge bases.
In this work, to solve (1), we propose an efficient $O(n)$ abduction mechanism based on KBC (\S\ref{data_creation}).
For (2), we develop {\tt abduction} tactic, that enables adding new lexical axioms
without quitting a Coq process~(\S\ref{sec:abduction_tactic}).

\begin{table*}[t]
    \centering
     \scalebox{0.95}{
    \begin{tabular}{ccl} \hline
        Relation $r$ & Generated Axiom & \multicolumn{1}{c}{Example} \\ \hline
        {\tt synonym}, {\tt hypernym}, & \multirow{2}{*}{$\forall x. s (x) \to o (x)$} & \multirow{2}{*}{$({\tt make}, {\tt synonym}, {\tt build}) ~\leadsto \forall e. {\tt make}(e) \to {\tt build}(e)$} \\
        {\tt derivationally-related} &  &  \\ \hdashline
        {\tt antonym}           & $\forall x.  s (x) \to \lnot o (x)$ &
        $({\tt parent}, r, {\tt child}) ~\leadsto \forall x. {\tt parent}(x) \to \lnot {\tt child}(x)$ \\ \hdashline
        {\tt hyponym}           & $\forall x. o (x) \to s (x)$ &
        $({\tt talk}, r, {\tt advise}) ~\leadsto \forall e. {\tt advise}(e) \to {\tt talk}(e)$ \\ \hline
    \end{tabular}
     }
    \caption{
        Triplet $(s,r,o)$ and axioms generated in terms of $r$.
        The type of an argument is determined in the semantic parsing part of ccg2lambda.
        While we use unary predicates (Neo-Davidsonian semantics), it can be generalized to any arity.
    }
    \label{generated_axioms}
\end{table*}

\begin{figure*}[t]
    \centering
    \includegraphics[scale=0.5]{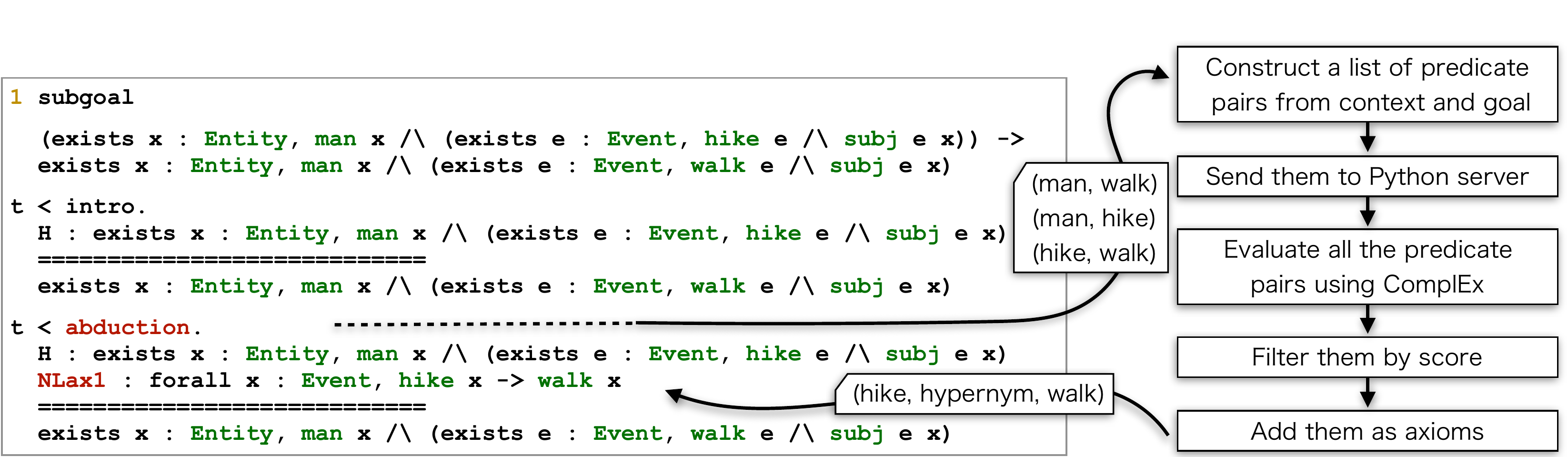}
    \caption{Running example of {\tt abduction} tactic in a Coq session proving
        ``{\it A man hikes}'' $\to$ ``{\it A man walks}''.
        When the tactic is executed, it interacts with a ComplEx model on a different process
        and injects high scoring triplets as axioms.
    }
    \label{abduction_tactic}
\end{figure*}

\section{Proposed method}
\label{sec:proposed}
Our abduction mechanism adds new axioms whenever the prover stops due to the lack of lexical knowledge.
Specifically, the system collects pairs of predicates from a current proof state and evaluates the pairs
for every relation $r \in \mathcal{R} = \{\mathtt{synonym}, \mathtt{hypernym}, \mathtt{antonym}, \mathtt{hyponym},$
$\mathtt{derivationally\mbox{--}related}\}$ with ComplEx~(Eq.~\ref{complex}).
It then declares as axioms logical formulas converted from the
triplets whose scores are above a predefined threshold $\theta$.
See Table~\ref{generated_axioms} for the conversion rules.
We describe the construction of a training data for ComplEx in \S\ref{data_creation}
and {\tt abduction} tactic that performs the axiom injection in \S\ref{sec:abduction_tactic}.

\subsection{Data creation}
\label{data_creation}
Although there already exist benchmark datasets for WordNet completion (e.g., WN18~\cite{Bordes:2013:TEM:2999792.2999923}), we construct our own dataset.
We find two problems on the existing datasets considering its application to RTE tasks.
One is a gap between the entities and relations appearing in those benchmark datasets and those needed to solve RTE datasets.
For example the knowledge on disease names are not necessary for the present dataset.

Another, more critical issue is that in WordNet many relations including hypernym and hyponym are defined among {\it synsets}, i.e., sets of synonymous lemmas, and the existing datasets also define a graph on them.
This is problematic in practice for ccg2lambda, which normalizes a word's surface form into its lemma when obtaining a logical formula.
A possible option to mitigate this discrepancy is to normalize each word into synset by adding word-sense disambiguation (WSD) step in ccg2lambda.
Another option is to construct a new dataset in which each relation is defined among lemmas.
We choose the latter for two reasons:
(1) existing WSD systems do not perform well and cause error propagation;
and (2) a dataset defined among lemmas eases the data augmentation using other resources.
For example, VerbOcean defines relations between lemmas so it is straightforward to augment the training data with it, as we show later.

We use different strategies to extract relations for each relation $r \in \mathcal{R}$ from WordNet as follows.\footnote{
    For simplicity, we use the set theoretical notation:
    $l \in s$ denotes lemma $l$ is an element of synset $s$.
}

\begin{description}
\item[synonym]
Since synonym relation is not defined in WordNet, we find them from other relations.
We regard two synsets $s_1$ and $s_2$ as synonym if they are in {\tt also\_sees}, {\tt verb\_groups}, or {\tt similar\_tos} relation, or if they share some lemma $l \in s_1 \cap s_2$.
Then, we take a Cartesian product of $s_1$ and $s_2$, that is, 
$(l_1, {\tt synonym}, l_2)$ for all $l_1 \in s_1$ and $l_2 \in s_2$, and add them to the dataset.
\item[hypernym and hyponym]
These relations are defined among synsets in WordNet.
As in {\tt synonym}, for {\tt hypernym} we take a Cartesion product of $s_1$ and $s_2$, when they are in hypernym relation.
We also collect the triplets obtained from its transitive closure,
since {\tt hypernym} is a transitive relation.\footnote{e.g., $({\tt puppy}, {\tt hypernym}, {\tt animal})$ follows from $({\tt puppy}, {\tt hypernym}, {\tt dog})$
and $({\tt dog}, {\tt hypernym}, {\tt animal})$.}
We process hyponym relations in the same way.
\item[antonym and derivationally-related]
Since antonym and derivationally-related relations are defined among lemmas in WordNet,
we simply collect them.
\end{description}

Since the constructed dataset contains many entities that will not be used
in the existing RTE datasets,
we strip off triplets $(s,r,o)$ if $s$ or $o$ is not found in our predefined lemma list, consisting of all lemmas of words appearing in the training part of
SICK~\cite{MARELLI14.363.L14-1314} and SNLI~\cite{snli:emnlp2015},
as well as the pre-trained GloVe word vectors.\footnote{
    \url{https://nlp.stanford.edu/projects/glove/}.
    We use the one with 6 billion tokens.
}
The resulting dataset contains 1,656,021 triplets and the number of
the entities is 41,577.
We spare random 10,000 triplets for development and use the rest for training.

VerbOcean is a semi-automatically constructed repository of semantic relations among verbs.
Since it defines a triplet among lemmas, we simply map the relations in the dataset to $\mathcal{R}$ after filtering triplets using the lemma list above.
Concretely, in the experiments we use {\tt similar} relations, which consists of 17,694 triplets.

\subsection{Axiom injection with {\tt abduction} tactic}
\label{sec:abduction_tactic}

As mentioned earlier, the abduction method in the previous work needs
to rerun Coq when adding new axioms.
Now we introduce our {\tt abduction} tactic that enables adding
new axioms on the fly without quitting the program.

Figure~\ref{abduction_tactic} shows a running example of {\tt abduction} tactic.
When executed, this tactic collects from the context~(the set of hypotheses) and
the goal~(the theorem to be proven) all the pairs of predicates and sends them to
a Python server running on other process.
The Python program evaluates the pairs for all relations in $\mathcal{R}$ with ComplEx and
then sends back the triplets whose score is above a threshold $\theta$.
In Coq, these are converted to logical formulas according to Table~\ref{generated_axioms}
and finally added to the context as axioms.

In Coq, as one can compose tactics into another tactic, it is possible to
construct a tactic that performs higher-order reasoning 
for natural language
that additionally injects lexical axioms when needed.
Note that search-based axiom injection also can be performed with this tactic,
by replacing the KBC-based scoring function with the search on databases.

We should note that one can combine our {\tt abduction} tactic with other semantic theories (e.g.~\citealp{chatzikyriakidis2014natural}; \citealp{bernardy17}) and put the system in application.
Coq has been used mainly for system verification and formalization of mathematics and there has been no tactic that is solely aimed at natural language reasoning. However, Coq provides an expressive formal language that combines higher-order logic and richly-typed functional programming language, and thus offers a general platform for various natural language semantics.
We believe that with our work, it will be easier to develop an NLP systems based on advanced linguistic theories.
This architecture also opens up new opportunities to integrate theorem proving and sophisticated machine-learning techniques.
For example, we could implement in a tactic more complex tasks such as premise selection with deep models~\cite{45402}.

\section{Experiments}
\label{experiment}

\subsection{SICK dataset}
We evaluate the proposed method on SICK dataset~\cite{MARELLI14.363.L14-1314}.
The dataset contains 4,500 problems (a pair of premise and hypothesis) for training, 500 for trial and 4,927 for testing, with a ratio of {\it entailment} / {\it contradiction} / {\it unknown} problems of .29 / .15 / .56 in all splits.
Note that ccg2lambda itself is an unsupervised system and does not use any training data.
We use the train part for the lemma list in data construction~(\S\ref{data_creation}) only and use the trial part to determine a threshold $\theta$ and to evaluate the processing speed.

\begin{table}[t]
    \centering
    \scalebox{0.85}{
    \begin{tabular}{l} \hline
        Id: (a) \ Label: \textsc{contradiction}  \\ 
       P: {\it A white and tan dog is running through the tall and green grass} \\
       H: {\it A white and tan dog is ambling through a field} \\ \hdashline
      Id: (b) \ Label: \textsc{entailment} \\
      P: {\it Someone is dropping the meat into a pan} \\
      H: {\it The meat is being thrown into a pan} \\ \hdashline
      Id: (c) \ Label: \textsc{entailment} \\
      P: {\it The man is singing and playing the guitar} \\
      H: {\it The guitar is being performed by a man} \\ \hline
      Id: (d) \ Label: \textsc{contradiction} \\
     P: {\it A man and a woman are walking together through the wood} \\
     H: {\it A man and a woman are staying together} \\ \hdashline
       Id: (e) \ Label: \textsc{entailment} \\
      P: {\it A man is emptying a container made of plastic completely} \\
      H: {\it A man is clearing a container made of plastic completely} \\ \hdashline
\end{tabular}}

\caption{
    LexSICK RTE problems
    require a mix of logical reasoning and
    external lexical knowledge.
    }
\label{lexsick_examples}
\end{table}

\subsection{New LexSICK lexical inference dataset}
While SICK dataset provides a good testbed for evaluating logical inference involving
linguistic phenomena such as negation and quantification,
we found that the dataset is not ideal for evaluating complex lexical inference,
specifically latent knowledge learned by a KBC model.

To evaluate the capability of our knowledge-empowered logic-based method,
we construct our own dataset, which is small
but challenging because of its combination of non-trivial lexical gaps and linguistic phenomena.
Table~\ref{lexsick_examples} shows example problems,
where lexical inferences are combined with linguistic phenomena such as
quantification (Example b), verb coordination (Example c), and passive-active alternation (Example b, c).
The process of the dataset construction is as follow:
a human expert picks a sentence (premise) from SICK dataset, changes a word to its synonym/antonym according
to thesauruses,\footnote{
  We avoided WordNet and used
  thesaurus.com~(\url{http://www.thesaurus.com/}) and
  Merriam-Webster~(\url{http://www.merriam-webster.com/}).
}
and then changes its sentence structure as well to construct a hypothesis.

The dataset~(we refer to it as LexSICK) contains 58 problems,
29 of which is labeled {\it entailment}, 29 {\it contradiction}, and no {\it unknown} label.
The whole dataset is publicly available.\textsuperscript{\ref{codebase}}

\subsection{Experimental settings}
\paragraph{Settings for ComplEx}
Unless otherwise stated, we set the dimension of embeddings $n = 50$ and train it on the triplets obtained from WordNet (excluding VerbOcean triplets) in \S\ref{data_creation}
by minimizing the logistic loss~(Eq. \ref{logistic}) using Adam optimizer.
We use 1-N scoring~(\S\ref{sec:kbc}), since our dataset is fairly large compared to standard benchmark datasets.
For the other hyperparameters, we use the same setting as in \citet{trouillon2016complex}, except for the batch size of 128.
In Table~\ref{kbc_results}, we show Mean Reciprocal Rank~(MRR) and Hits@$N$~($N=1,3,10$) of ComplEx model
~(and the state-of-the-art ConvE~\cite{DBLP:journals/corr/DettmersMSR17} with the default hyperparameters for comparison) on development part of the dataset in \S\ref{data_creation}.\footnote{
\label{filtered}
We report the scores in filtered setting. That is, compute the
rank of $o$ for gold $(s, r, o)$ against all $e \in \mathcal{E}$ such that $(s, r, e)$ is not in
either of training or development data.
}
The ComplEx model scores 77.68 in MRR, which is comparably lower than scores reported for the widely used WN18 benchmark data~(above 93).
Notably, ComplEx performs better than ConvE in terms of all metrics in this experiment.
We adopt ComplEx in this work, since it achieves better results with much lower computational load.

\begin{table}[t]
    \centering
    \begin{tabular}{cccccc} \hline
        \multirow{2}{*}{\bf Method} &  \multirow{2}{*}{\bf MRR} & \multicolumn{3}{c}{\bf Hits}  \\ \cline{3-5}
         &  & @1 & @3  & @10 \\ \hline
        ComplEx & 77.68 & 71.07 & 81.76 & 90.08 \\
        ConvE   & 67.41 & 57.11 & 75.02 & 85.76 \\ \hline
    \end{tabular}
    \caption{
        The performance of KBC models trained and evaluated on WordNet triplets in \S\ref{data_creation}.
        We report filtered scores\textsuperscript{\ref{filtered}}.
}
    \label{kbc_results}
\end{table}

\paragraph{Settings for ccg2lambda}
We decide the threshold $\theta=0.4$ for filtering triplets
based on the accuracy on the SICK trial set.
As baselines, we replicate the system of \citet{martinezgomez-EtAl:2017:EACLlong} 
with the default settings of ccg2lambda\footnote{
	We use a version of ccg2lambda committed to the master branch on October 2, 2017.
} except for the use of CCG parsers mentioned below.

\begin{table*}[t]
    \centering
    \begin{tabular}{cccccccccccc} \hline
        \multirow{2}{*}{\bf System} & \multicolumn{3}{c}{\bf Method} & \multicolumn{2}{c}{\bf Dataset} & \multirow{2}{*}{\bf Accuracy} & \multirow{2}{*}{\bf Precision}& \multirow{2}{*}{\bf Recall} & \multirow{2}{*}{\bf F1}  & \multirow{2}{*}{\bf Speed} \\ \cline{2-6}
        & search & KBC & tactic & WN & VO & & & & & \\ \hline
        \citet{mineshima-EtAl:2015:EMNLP} & & & & & & 77.30 &  98.93 & 48,07 & 64.68 & 3.79 \\
        \multirow{2}{*}{\citet{martinezgomez-EtAl:2017:EACLlong}} & \cmark & & & \cmark & & 83.55 & 97.20 & 63.63 & 76.90 & 9.15 \\
        & \cmark & & & \cmark & \cmark & 83.68 &  96.88 & 64.15 & 77.16 & 9.42 \\ \hdashline
        \multirow{3}{*}{Ours} & \cmark & & \cmark & \cmark & \cmark & 83.64 & 97.15 & 64.01 & 77.16 & 7.07 \\
        & & \cmark & \cmark & \cmark & & 83.55 &  96.28 & 64.38 & 77.14 & 4.03 \\
        & & \cmark & \cmark & \cmark & \cmark & 83.45 &  95.75 & 64.47 & 77.04 & 3.84 \\ \hline
    \end{tabular}
    \caption{
        Results on SICK test set. The results of the baseline systems are above the dashed line.
        The {\bf Method} columns represent the use of search-based abduction,
        KBC-based abduction and {\tt abduction} tactic,
        while the {\bf Dataset} columns datasets used in abduction: WordNet~(WN) and VerbOcean~(VO).
       {\bf Speed} shows macro average of processing time (sec.) of an RTE problem.
}
    \label{sick_result}
\end{table*}

\begin{table}
    \centering
    \begin{tabular}{cc} \hline
        {\bf Method} & {\bf \#Correct}  \\ \hline
       ResEncoder~\cite{nie2017shortcut} &  18~/~58 \\ \hdashline
        search-based abduction & 20~/~58 \\
        KBC-based abduction & 21~/~58  \\ \hline
    \end{tabular}
    \caption{ Experimental results on LexSICK. 
    Both search- and KBC-based abductions use WordNet and VerbOcean.}
    \label{lexsick_result}
\end{table}

One bottleneck of ccg2lambda is errors in CCG parsing.
To mitigate the error propagation, it uses four CCG parsers and aggregates the results:
C\&C~\cite{J07-4004}, EasyCCG~\cite{lewis-steedman:2014:EMNLP2014},
EasySRL~\cite{D15-1169} and depccg~\cite{yoshikawa-noji-matsumoto:2017:Long}.
We regard two results as contradicted with each other if one is {\it entailment} and
the other is {\it contradiction}.
In such cases the system outputs {\it unknown};
otherwise, if at least one parser results in {\it entailment} or {\it contradiction},
that is adopted as the system output.

We report accuracy / precision / recall / F1 on the test part and
processing speed (macro average of five runs) in the trial set.\footnote{
    We preprocess each RTE problem and do not include in the reported
    times those involved with CCG/semantic parsing.
    We set the time limit of proving to 100 seconds.
    These experiments are conducted on a machine with 18 core 2.30 GHz Intel Xeon CPU~$\times$~2. }

\subsection{Results on SICK set}

Table~\ref{sick_result} shows the experimental results on SICK.
The first row is a baseline result without any abduction mechanism, followed by ones using
WordNet and VerbOcean additively with the search-based axiom injection.
By introducing our {\tt abduction} tactic that is
combined with the search-based axiom injection~(4th row),
we have successfully reduced the computation time~(-2.35 sec. compared to 3rd row).
By replacing the search-based abduction with the ComplEx model,
the averaged time to process an RTE problem is again significantly reduced~(5th row).
The time gets close to the baseline without any database~(only +0.24 sec.), with much improvement in terms of RTE performance, 
achieving the exactly same accuracy with \citeauthor{martinezgomez-EtAl:2017:EACLlong}'s WordNet-based abduction.

Finally, we re-train a ComplEx model with {\tt similar} relations from VerbOcean~(final row).
When combining the datasets, we find that setting the dimension of embeddings larger to $n = 100$
leads to better performance. This may help the KBC model accommodate the relatively noisy nature of VerbOcean triplets.
The VerbOcean triplets have contributed to the improved recall by providing more knowledge not covered by WordNet,
while it has resulted in drops in the other metrics.
Inspecting the details, it has actually solved more examples than when using only WordNet triplets;
however, noisy relations in the data, such as $({\tt black}, {\tt similar}, {\tt white})$,
falsely lead to proving {\it entailment}~/~{\it contradiction} for problems whose true label is {\it unknown}.
The higher recall has contributed to the overall processing speed,
since it has led more problems to be proven before the timeout~(-0.25 sec.).

We conduct detailed error analysis on 500 SICK trial problems.
The RTE system combined with our KBC-based abduction (trained only on WordNet) has correctly solved one more problem than a \citeauthor{martinezgomez-EtAl:2017:EACLlong}'s baseline system (which additionally uses VerbOcean), resulting in the accuracy of 84.8\%.
In this subset, we found the following error cases:
71 cases related to lexical knowledge, 4 failures in CCG parsing, and 1 timeout in theorem proving.
This shows that CCG parsing is quite reliable. Around five cases among lexical ones are due to false positively asserted lexical axioms, while the most of the others are due to the lack of knowledge.
In the following, we exclusively focus on issues related to lexical inference.

\subsection{Evaluating latent knowledge}
Table~\ref{lexsick_result} shows experimental results on LexSICK dataset.
For comparison, we add a result of ResEncoder~\cite{nie2017shortcut},
one of the state-of-the-art neural network-based RTE models.
When trained on the training part of SICK,
it scores $82.00\%$ accuracy on the SICK test part,
while performs badly on LexSICK,
indicating this dataset is more challenging.
The accuracies of two logic-based methods are also not high, suggesting the difficulty of this problem.
The KBC-based method shows the same tendency as
in the results in the previous section;
it solves more examples than the search-based method,
along with false positive predictions.

We observe interesting examples that show the effectiveness of using latent lexical knowledge.
One is Example (d) in Table~\ref{lexsick_examples}, for which
a latent relation $({\tt walk}, {\tt antonym}, {\tt stay})$ is predicted by the
KBC model, while it is not available for the search-based method.
Another case is Example (e) in Table~\ref{lexsick_examples},
where our method obtains the axiom
$\forall x. \mathtt{clear}(x) \tto \mathtt{empty}(x)$
by scoring the triplet $({\tt empty}, {\tt synonym}, {\tt clear})$
as high as $({\tt empty}, {\tt hyponym}, {\tt clear})$,
leading to the correct prediction.
The search-based method derives only
$\forall x. \mathtt{empty}(x) \tto \mathtt{clear}(x)$,
which is not relevant in this case.
     
Similarly, by examining the results on SICK dataset,
we found more examples that the KBC-based method has successfully solved by utilizing latent lexical knowledge.
For example, in \textsc{SICK}-6874
we have a premise {\it A couple of white dogs are running and jumping along a beach}
and a hypothesis {\it Two dogs are playing on the beach}.
The KBC model successfully proves the inference by producing multiple axioms:
$\forall x. {\tt along}(x) \tto {\tt on}(x)$,
$\forall x. {\tt couple}(x) \tto {\tt two}(x)$ and
$\forall x. {\tt run}(x) \tto {\tt play}(x)$.
One characteristic of the KBC-based method is that it can utilize
lexical relations between general words such as frequent verbs and prepositions.
Though this may cause more false positive predictions,
our experiments showed that under the control of the abduction method,
it contributed to improving recall while keeping high precision.

\section{Conclusion and future direction}
In this work, we have proposed an automatic axiom injection mechanism for natural language reasoning based on Knowledge Base Completion.
In the experiments, we show that our method has significantly improved the processing speed of an RTE problem,
while it also achieves the competitive accuracy, by utilizing the obtained latent knowledge.

In future work, we are interested in extending this framework to generate phrasal axioms~(e.g., $\forall x. {\tt have}(x) \land {\tt fun}(x) \to {\tt enjoy}(x)$).
Generating this sort of axioms accurately is the key for logic-based systems to achieve high performance.
In order to do that, we will need a framework to learn to compute
compositionally a vector from those of {\it have} and {\it fun}
and to predict relations such as {\tt synonym} between the vector and that of {\it enjoy}.

\section*{Acknowledgements}
We thank the three anonymous reviewers for their insightful comments.
We are also grateful to Bevan Johns for proofreading examples in LexSICK dataset
and Hitoshi Manabe for his public codebase from which we learned many about the KBC techniques.
This work was supported by JSPS KAKENHI Grant Number JP18J12945,
and also by JST CREST Grant Number JPMJCR1301.

\bibliographystyle{aaai}
\bibliography{yoshikawa_bib}

\end{document}